\documentclass[english]{article}
\usepackage{float}
\usepackage{graphics} % for pdf, bitmapped graphics files
\usepackage{epsfig} % for postscript graphics files
\usepackage{mathptmx} % assumes new font selection scheme installed
\usepackage{times} % assumes new font selection scheme installed
\usepackage{amsmath} % assumes amsmath package installed
\usepackage{amssymb}  % assumes amsmath package installed
\usepackage{bm}        % bold math
\usepackage{xcolor}
\usepackage{multirow}
\usepackage{array}
\usepackage{geometry}
\usepackage[round]{natbib}
\usepackage{placeins}
\usepackage{amsmath,amssymb}
\usepackage{algorithm}
\usepackage{algpseudocode}
\usepackage{float} % for [H] if needed
\usepackage{booktabs}
\usepackage{tabularx}
\usepackage{makecell}
\usepackage{enumitem}
\usepackage{placeins}
\usepackage{titling}
\usepackage{subcaption}
\usepackage{adjustbox}
\usepackage{url}
\usepackage[colorlinks=true,
            linkcolor=blue,
            citecolor=blue,
            urlcolor=blue]{hyperref}
\newcommand{\algcomment}[1]{\textit{// #1}}

\definecolor{own_pink}{RGB}{217,25,169}
\definecolor{own_blue}{RGB}{0,100,223}

\newcommand{\br}{\bm{r}}

\newcommand{\cB}{\mathcal{B}}

\newcommand{\cD}{\mathcal{D}}

\usepackage{scalerel,stackengine}
\stackMath
\newcommand\reallywidehat[1]{%
\savestack{\tmpbox}{\stretchto{%
  \scaleto{%
    \scalerel*[\widthof{\ensuremath{#1}}]{\kern-.6pt\bigwedge\kern-.6pt}%
    {\rule[-\textheight/2]{1ex}{\textheight}}%WIDTH-LIMITED BIG WEDGE
  }{\textheight}% 
}{0.5ex}}%
\stackon[1pt]{#1}{\tmpbox}%
}
\newcommand\reallywidecheck[1]{%
\savestack{\tmpbox}{\stretchto{%
  \scaleto{%zh
    \scalerel*[\widthof{\ensuremath{#1}}]{\kern-.6pt\bigwedge\kern-.6pt}%
    {\rule[-\textheight/2]{1ex}{\textheight}}%WIDTH-LIMITED BIG WEDGE
  }{\textheight}% 
}{0.5ex}}%
\stackon[1pt]{#1}{\scalebox{-1}{\tmpbox}}%
}

\title{\LARGE \bf
T2S-MPC: Time-Embedded Online Adaptive Model Predictive Control for Time-Varying Dynamics
}
\author{
Zeyu Shen\thanks{Department of Applied Mathematics and Statistics, Johns Hopkins University, MD, USA.}\\
JHU
\and
Zhuoyuan Wang\thanks{Department of Electrical and Computer Engineering, Carnegie Mellon University, PA, USA.}\\
CMU
\and
Laixi Shi\thanks{Department of Electrical and Computer Engineering, Johns Hopkins University, MD, USA. Corresponding author.}\\
JHU
}

\begin{document}
\maketitle

\begin{abstract}
Recent advances in learning-based model predictive control (MPC) have leveraged neural networks for online model learning, achieving strong performance when nonstationary system dynamics deviate from nominal models. However, existing approaches primarily address specific or relatively structured forms of dynamical variation, leaving more general, unknown, and unpredictable time-varying dynamics insufficiently handled. To tackle this challenge, we propose T2S-MPC, a framework that adaptively learns a residual dynamics model online and integrates it with the nominal model within the MPC framework to enable fast-evolving online planning.  To make the model time-aware, we explicitly encode temporal information through a structured time embedding and employ a two-timescale update scheme, allowing the controller to capture nonstationary dynamics while balancing rapid adaptation with stable learning. We evaluate the proposed method on a 2D quadrotor across stabilization and trajectory tracking tasks under diverse time-varying disturbances, including linear drifting and periodic perturbations. Experimental results show that T2S-MPC consistently outperforms classical MPC, neural MPC, and ablated variants in control performance, while also demonstrating strong robustness across a wide range of disturbance conditions without additional tuning. The source code is publicly available at:\noindent\textit{
\url{https://github.com/Zeyuu0920/T2S_MPC}
}
 
\end{abstract}

\noindent \textbf{Keywords:} machine learning for control, online learning, model predictive control, nonstationarity, adaptive control.

%!TEX root = ./DRO-Simulator.tex
\section{Introduction}

 Many real-world applications inherently involve nonstationary dynamics~\citep{patil2022adaptive,zhang2024global,shi2022prescribed}, where autonomous control frameworks increasingly taking on a central role. Among control frameworks, model predictive control (MPC) is an optimization-based control framework that plans control actions using a predictive model of system dynamics over a finite horizon, and is widely used in industrial process control, robotics, and other autonomous systems~\citep{garcia1989model, kouvaritakis2016model}. Recently, learning-based model predictive control (MPC) has emerged as a powerful control framework for handling more complex and nonstationary tasks~\citep{hewing2020learning,kabzan2019learning,ren2022tutorial}. Meanwhile, neural networks have become increasingly attractive for dynamics modeling due to their flexibility and expressive power in capturing complex nonlinear behaviors that are difficult to model analytically~\citep{kumpati1990identification, hunt1992neural}. Motivated by these advances, a recent line of work has begun to incorporate neural networks for online model learning directly within the MPC framework.

Existing approaches have explored different neural network architectures and have shown that neural online model learning can substantially improve control performance when the deployed system deviates from its nominal dynamics~\citep{Wu2019RealTimeMLMPC,jiahao2023online,mei2025fast}. However, those neural-network-based online-learning MPC framework still primarily focus on specific and relatively structured forms of time variation, such as static model mismatch~\citep{jiahao2023online}, bounded disturbances~\citep{Wu2019RealTimeMLMPC}, or adaptation across related operating conditions through meta-learning~\citep{mei2025fast}. While many practical systems exhibit more complex and diverse forms of nonstationarity, in which the dynamics may drift over time in unknown, less structured, and unpredictable ways. Examples include aerial and ground robots operating under changing wind~\citep{o2022neural}, terrain~\citep{achterhold2024learning}, or payload conditions~\citep{hanover2021performance}, as well as safety-critical systems affected by actuator degradation due to thermal effects, wear, or evolving external disturbances\citep{langeron2015modeling}.

\begin{figure}[t]
    \centering
    \includegraphics[width=0.6\linewidth]{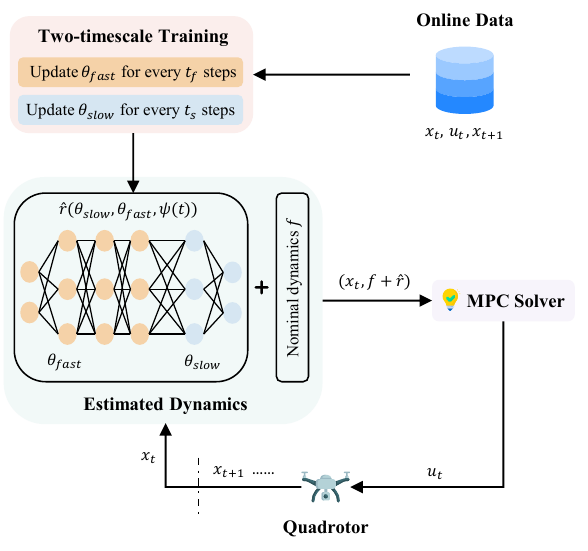}
    \caption{Overall diagram of the proposed T2S-MPC method.}
    \label{fig:overall_diagram}
\end{figure}

This motivates online model learning that can continuously adapt the predictive model during deployment to address broader forms of complex and unpredictable time-varying dynamics. In this work, we develop a online learning-based MPC framework that learns a neural-network-based time-aware residual dynamic model to compensate for the adaptive model mismatch. The main contributions of this work are summarized as follows:

\begin{itemize}[itemsep=0.2em]
\item We introduce Time-Embedded Two-Timescale Neural MPC (T2S-MPC), a online MPC framework that learns a residual model by incorporating a structured time embedding as an explicit input to the neural network, along with a two-timescale parameter update scheme. It enables fast adaptation to time-varying dynamics while maintaining stable learning over diverse disturbance patterns within a unified model. The overall diagram of the proposed method is summarized in Algorithm~\ref{alg:online_adaptive_mpc} and shown in Fig.~\ref{fig:overall_diagram}.

\item We evaluate T2S-MPC on quadrotor stabilization and trajectory tracking tasks under multiple time-varying disturbances, where it consistently achieves superior adaptation and control performance compared to classical MPC and neural MPC baselines (Table~\ref{tab:ablation_study}, Fig.~\ref{fig:stabilization_error_plot}-\ref{fig:tracking}). Extensive ablations across a wide range of disturbance magnitudes, frequencies, and patterns further demonstrate its consistent advantage over all baselines (Table~\ref{tab:style}).

\end{itemize}

\section{Related Work}

\paragraph{Learning-Based Adaptive Control for Dynamics Uncertainty.}

Classical adaptive control~\citep{aastrom1995adaptive, ioannou1996robust, chen2021adaptive} typically assumes parametric uncertainty with known dynamics structure, limiting its applicability to complex or general unstructured time-varying dynamics. Learning-based adaptive control addresses this through several important directions. One line embeds online-updated learned dynamics or residual models into classical control frameworks~\citep{wu2019real, sun2021online, jiahao2023online, mei2025fast,chakrabarty2023meta, muthirayan2025meta}. A second leverages meta-learning to acquire features, representations, or initializations that enable rapid adaptation across related environments~\citep{richards2021adaptive, shi2021meta, o2022neural, yan2024mpc}, involving offline and possibly also online learning processes~\citep{jiahao2023online,sun2021online,smith2020online,williams2019locally,gradu2023adaptive}. Some other directions emphasize stability and performance guarantees, for example by estimating parameters of structured (often linearly parameterized) uncertainty models~\cite{chowdhary2010concurrent, chowdhary2014bayesian, parikh2019integral, glushchenko2021drem, goel2024composite}, or by formulating control as regret minimization with provable sublinear regret~\cite{minasyan2021online, gradu2023adaptive}. Targeting general unstructured and unpredictable time-varying dynamics, we follow the first direction and learn a residual model within an MPC framework.

\paragraph{Learning-Based MPC.}
Within the line of integrating learned models into classical control frameworks, model predictive control (MPC) has emerged as a particularly effective framework~\citep{hewing2020learning, ren2022tutorial}; see \citet{wu2025tutorial} for a recent review. Among non-neural network approaches, Gaussian processes and related Bayesian nonparametric models are widely used to learn unknown dynamics or residual errors while quantifying uncertainty~\citep{wang2024tutorial, chowdhary2014bayesian}. More recently, neural network architectures—including recurrent networks~\citep{wu2019real} and neural ODE-based models—have been adopted for learning dynamics in MPC, with tools such as L4CasADi~\citep{salzmann2024learning} enabling seamless integration of neural dynamics into MPC pipelines. In robotics, nominal physics models are commonly augmented with neural residual components to improve prediction and control in legged locomotion and aerial systems~\citep{sun2021online, chee2022knode, jiahao2023online, salzmann2023real}. A parallel line incorporates meta-learning to acquire initializations, features, or adaptation mechanisms from offline task distributions, enabling rapid adaptation with limited online data at deployment~\citep{richards2021adaptive, shi2021meta, o2022neural, yan2024mpc, mei2025fast}.

Overall, existing learning-based MPC methods either rely on non-neural network probabilistic models, employ neural architectures that capture time variation only implicitly through online updates, or require offline meta-learning across related environments. In this work, we propose T2S-MPC, which explicitly models nonstationarity via time embeddings in the neural residual dynamics and couples this time-aware model with a two-timescale online update mechanism, enabling both fast adaptation and stable learning across diverse time-varying disturbances within a unified MPC framework.

\section{Problem Formulation}

We consider the adaptive control problem of a time-varying nonlinear dynamical system, with continuous-time dynamics at time $t$ defined as
\begin{equation}
\dot{\mathbf{x}}(t) = f_t(\mathbf{x}(t), \mathbf{u}(t)),
\end{equation}
where $\mathbf{x}(t) \in \mathbb{R}^n$ denotes the system state, $\mathbf{u}(t) \in \mathbb{R}^m$ is the control input, and $f_t(\cdot)$ represents the time-varying system dynamics at time $t$.

In many practical scenarios, the system dynamics are not fully known, and only a partially known approximate nominal model is available. Therefore, we model the system dynamics as the combination of a nominal model and a time-varying residual term~\citep{mei2025fast,saveriano2017data}, expressed as follows, 
\begin{equation}\label{eq:model-decomposition}
f_t(\mathbf{x}(t), \mathbf{u}(t)) = f(\mathbf{x}(t), \mathbf{u}(t)) + \br_t(\mathbf{x}(t), \mathbf{u}(t)).
\end{equation}
Here, $f : \mathbb{R}^n \times \mathbb{R}^m \rightarrow \mathbb{R}^n$ represents the known approximate nominal dynamics, and $\mathbf{r}_t : \mathbb{R}^n \times \mathbb{R}^m \rightarrow \mathbb{R}^n$ captures the unknown time-varying residual dynamics at time $t$, i.e., the discrepancy between the nominal model $f(\cdot)$ and the true dynamics $f_t(\cdot)$. Time-varying dynamics commonly arise in practice due to external disturbances and modeling uncertainties. For instance, in robotic systems, one can consider $\mathbf{r}_t(\mathbf{x}(t), \mathbf{u}(t)) = \mathbf{w}(t)$ being the state- and action-independent stochastic noise.

At the current time $t$, the goal of the controller is to determine the optimal control input $\mathbf{u}(t)$ that minimizes a future long-term cost. This leads to the following optimal control problem (OCP):
\begin{equation}
\begin{aligned}\label{eq:continuous-optimal-conrtol}
\min_{\mathbf{u}(\cdot)} \quad &  \int_{\tau = t}^{t+t_p} \ell(\mathbf{x}(\tau), \mathbf{u}(\tau)) \, d\tau  + m(\mathbf{x}(t+t_p))  \\
\text{s.t.} \quad 
& \dot{\mathbf{x}}(\tau) = f_t(\mathbf{x}(\tau), \mathbf{u}(\tau)), \\
& \mathbf{c}(\mathbf{x}(\tau), \mathbf{u}(\tau)) \leq 0.
\end{aligned}
\end{equation}
Here, $t_p$ denotes the online prediction horizon, $\ell(\mathbf{x}(\tau), \mathbf{u}(\tau))$ is the running cost incurred at time $\tau$, and $m(\mathbf{x}(t+t_p))$ is the terminal cost. The constraint $\mathbf{c}(\mathbf{x}(\tau), \mathbf{u}(\tau)) \le 0$ represents state and control constraints arising from safety requirements or environmental limitations. For example, $\mathbf{c}$ may encode box constraints on the control input, state bounds that keep the system within a safe operating region, or obstacle-avoidance constraints in robotic navigation.

This problem is particularly challenging because the controller must optimize future behavior over the horizon $[t, t+t_p]$ while the true dynamics $f_t = f + \mathbf{r}_t$ are only partially known and may vary over time. In particular, accurate prediction of future states depends on learning the unknown residual dynamics $\mathbf{r}_t(\mathbf{x}, \mathbf{u})$ online, while the control policy must simultaneously remain responsive to both persistent temporal variations and sudden changes in the dynamics. This creates an inherent tension between maintaining accurate predictions and achieving timely responsiveness under nonstationary conditions when solving the OCP.

\section{Time-Dependent Learning-Based MPC}

Neural networks have emerged as powerful function approximators for dynamical systems due to their strong expressive and generalization capabilities \citep{zhang2016understanding}. Inspired by this, we adopt the online adaptive neural model predictive control (MPC) framework~\citep{mei2025fast}, in which a neural network is used to model the potentially time-varying unknown residual dynamics in real time and is then integrated with the nominal model in the MPC solver for subsequent planning.

Learning-based MPC is an optimization-based approach to optimal control that 
employs a learned dynamical model for planning. At each time step, it solves 
a finite-horizon optimal control problem online given the current state and 
the learned model, then applies only the first control element of the resulting 
control sequence before re-solving at the next step. Mathematically, given the learned model $g_t$ at time $t$, MPC solves the 
following finite-horizon optimal control problem over a prediction horizon $N$:
\begin{equation}
\begin{aligned}
\min_{\mathbf{u}_{0:N-1}} \quad & \sum_{k=0}^{N-1} \ell(\mathbf{x}_k, \mathbf{u}_k) + m(\mathbf{x}_N) \\
\text{s.t.} \quad 
& \mathbf{x}_{k+1} = g_t(\mathbf{x}_k, \mathbf{u}_k), \quad k = 0, \ldots, N-1, \\
& \mathbf{x}_0 = \mathbf{x}(t), \\
& \mathbf{c}(\mathbf{x}_k, \mathbf{u}_k) \leq 0, \quad k = 0, \ldots, N-1,
\end{aligned}
\label{eq:mpc_discrete}
\end{equation}
where $N$ denotes the number of discretized steps in the prediction horizon. Note that~\eqref{eq:mpc_discrete} can be viewed as a discretized version 
of~\eqref{eq:continuous-optimal-conrtol}, with the learned model dynamics 
represented by the general input $g_t$. The next state $\mathbf{x}_{k+1} = 
g_t(\mathbf{x}_k, \mathbf{u}_k)$ can be computed via methods such as the 
direct multiple shooting method~\citep{bock1984multiple}, where a numerical 
integrator discretizes the continuous-time dynamics $g_t$ over one time step 
$\delta t = \frac{t_p}{N}$.

At each control time step $t$, MPC takes the current state $\mathbf{x}(t)$ 
and learned dynamics $g_t$, solves the optimal control problem~\eqref{eq:mpc_discrete} 
over horizon $N$, and obtains the optimal control sequence $\mathbf{u}_{0:N-1}^\star$. 
Only the first control input $\mathbf{u}^\star_0$ is then applied to the system.

\subsection{Neural Residual Dynamic Learning}\label{sec:residual-learning}

With MPC in place, the key challenge is that the time-varying system dynamics $f_t(\cdot)$ in~\eqref{eq:model-decomposition} are unknown, whereas MPC requires a known model as input. To address this, we propose to use a neural network to learn an online estimate $\hat{\mathbf{r}}_t(\cdot)$ of the time-varying residual dynamics $\mathbf{r}_t(\cdot)$. This estimate can then be combined with the known nominal model $f(\cdot)$ to form an approximate system dynamics model $\hat{f}_t(\cdot) = f(\cdot) + \hat{\mathbf{r}}_t(\cdot)$.
Accordingly, the remainder of this subsection focuses on learning an accurate neural residual model $\hat{\mathbf{r}}_t$ that approximates the ground-truth residual dynamics $\mathbf{r}_t$, which can then be integrated into the MPC framework for online adaptive control.

As the system of interest (e.g., a robot) is deployed online, a streaming sequence of data samples becomes available at each time $t$. Each sample consists of the current state, control input, and the next state, i.e., $\big[\mathbf{x}(t_i), \mathbf{u}(t_i), \mathbf{x}(t_{i+1}) \big]$. We construct the online dataset as $\mathcal{D}_t = \big\{ \big[\mathbf{x}(t_i), \mathbf{u}(t_i), \mathbf{x}(t_{i+1}) \big] \big\}_{i=1}^{D_t}$, where $D_t$ denotes the number of samples in $\mathcal{D}_t$.
For each sample at time $t_i$, the ground-truth residual can be computed from two consecutive states $\mathbf{x}(t_i)$ and $\mathbf{x}(t_{i+1})$ as
\begin{equation}
\forall i \in \{1,2,\cdots, D_t\}, \quad \mathbf{r}_{t_i}(\mathbf{x}(t_i), \mathbf{u}(t_i)) = \mathbf{x}(t_{i+1}) - \mathbf{x}(t_i).
\label{eq:residual_target}
\end{equation}
Neural residual models aims to approximate the ground-truth residual dynamics with a neural network. 

\subsection{Time-Embedded Two-Timescale Learning}

We introduce a time-embedded two-timescale learning method to approximate the ground-truth residual dynamics. The model takes the current state, control input, and time as inputs, and outputs the estimated residual. Specifically, at each control time $t_i$, the residual is estimated as
\begin{equation}
\hat{\mathbf{r}}_t(\mathbf{x}(t_i), \mathbf{u}(t_i)) = \hat{\mathbf{r}}_{\theta} (\mathbf{x}(t_i), \mathbf{u}(t_i), \psi(t_i)),
\label{eq:residual_modeling}
\end{equation}
where the neural residual model is parameterized by $\theta$, and $\psi(t)$ denotes a time embedding that is explicitly included as an input. To better learn the neural residual model for diverse time-varying dynamics online, the parameter $\theta$ is updated using online data $\mathcal{D}_t$, following a two-timescale learning process. The two key designs of the proposed methods is introduced below.

\paragraph{Time embedding}
\label{sec:Time-Aware Representation}

To capture the non-stationary nature of the dynamics, we incorporate time information into the residual model via a structured time embedding $\psi: \mathbb{R} \rightarrow \mathbb{R}^d$ that takes raw time as an input and returns a $d$-dimensional vector denoted as the time-embedding.
\begin{equation}
\label{eq:time_embedding}
\psi(t) = \Big[
\sin(\omega_1 t), \dots, \sin(\omega_{d/2} t),
\cos(\omega_1 t), \dots, \cos(\omega_{d/2} t)
\Big],
\end{equation}
where the frequencies are defined as
\begin{equation}
\omega_i = \frac{\pi}{i}, \quad i = 1, \dots, \frac{d}{2}.
\end{equation}
The time embedding $\psi(t)$ is concatenated with the state and control inputs $\big[ \mathbf{x}(t), \mathbf{u}(t), \psi(t) \big]$ forming the augmented input to the residual model. Note that our framework allows for flexible selection of potential time embedding forms, and the one considered in \eqref{eq:time_embedding} is one such example.

\begin{algorithm}[t]
    \caption{Time-Embedded
    Two-Timescale Neural MPC (T2S-MPC)}
    \label{alg:online_adaptive_mpc}
    \begin{algorithmic}[0.8]
    \State \textbf{Input:} Nominal model $f(\cdot)$, task horizon $T_\text{sim}$, MPC prediction horizon $N$, update periods $(t_f, t_s)$, batch sizes $(B_f, B_s)$.
    \State \textbf{Initialize:} $\theta = \{\theta_f, \theta_s\}$, state $\mathbf{x}(0)$, data buffer $\mathcal{D} = \{\}$
    
    \For{$t_i = 0, \delta t, 2\delta t, \dots, \delta(T_\text{sim}-1)$}
    
        \State \algcomment{Output control solution using MPC with current estimated dynamic model}
        \State Let model $g_t = f(\mathbf{x}(t_i), \mathbf{u}(t_i)) + \hat{\mathbf{r}}_{\theta} (\mathbf{x}(t_i), \mathbf{u}(t_i), \psi(t_i))$
        \State Plugging $g_t$ in MPC framework in \eqref{eq:mpc_discrete} and output the control solution sequence $\mathbf{u}_{0:N-1}^\star$
        \State Apply the first control input $\mathbf{u}(t_i) \gets \mathbf{u}_0^\star$ in the system and observe the next system state $\mathbf{x}(t_{i+1})$
        \State Insert tuple $\big[ \mathbf{x}(t_{i}), \mathbf{u}(t_{i}), \mathbf{x}(t_{i+1}) \big]$ into dataset $\mathcal{D}$.
    
        \State \algcomment{Update neural residual model $\theta = \{\theta_f, \theta_s\}$}
        
        \If{$t_i \bmod t_f = 0$}
            \State Get data batch $\cB_f$ as the latest $|\cB_f|$ samples from $\mathcal{D}$ 
            \State Calculate the true residual model in \eqref{eq:residual_target} and estimation in \eqref{eq:residual_modeling} for each sample in  $\cB_f$
            \State Update fast-updated parameters $\theta_f$ through \eqref{eq:fast_update}
        \EndIf
        
        \If{$t_i \bmod t_s = 0$}
            \State Sample data batch $\cB_s$ randomly from $\mathcal{D}$
            \State Calculate the true residual model in \eqref{eq:residual_target} and estimation in \eqref{eq:residual_modeling} for each sample in  $\cB_s$
            \State Update slow-updated parameters $\theta_s$ through \eqref{eq:slow_update}
        \EndIf
        
    \EndFor
    \end{algorithmic}
\end{algorithm}

\paragraph{Two-Timescale Online Learning}

To enable both rapid adaptation and stable learning, we adopt a two-timescale optimization scheme.  We split the parameters of the neural residual model into two groups, i.e., $\theta = \{\theta_f, \theta_s\}$.
\begin{itemize}
    \item Fast-updated $\theta_f$: updated frequently by every $t_f$ time steps to adapt to short-term changes, 
    \item Slow-updated $\theta_s$: updated every $t_s > t_f$ time steps, less frequently to maintain stability.
\end{itemize}

During the online learning process, at any time $t$, the loss function for any given batch size of data $\cB \in \cD_t$ is defined by the empirical error between ground truth residual model and the output of the neural residual model:
\begin{equation}
\mathcal{L}(\cB) = \frac{1}{|\cB|}\sum_{\cB} \left\| \mathbf{r}_t(\mathbf{x}(t_i), \mathbf{u}(t_i)) - \hat{\mathbf{r}}_{\theta} (\mathbf{x}(t_i), \mathbf{u}(t_i), \psi(t_i)) \right\|^2.
\label{eq:loss}
\end{equation} 

After each $t_f$ control steps, the fast-update parameters $\theta_f$ are updated using the learning rate $\eta_f$ as
\begin{equation}
\theta_f \leftarrow \theta_f - \eta_f \nabla_{\theta_f} \mathcal{L}(\cB_f) , \label{eq:fast_update} 
\end{equation}
where $\mathcal{B}_f$ is a batch constructed from the most recent $b_f$ samples.

For slowly updated parameters, after each $t_s$ ($t_s >t_f$) control step, the parameters $\theta_s$ are updated using the learning rate $\eta_s$ as
\begin{equation}
\theta_s \leftarrow \theta_s - \eta_s \nabla_{\theta_s} \mathcal{L}(\cB_s), \label{eq:slow_update}
\end{equation}
where $\cB_s$ is the data batch randomly sampled from the entire online dataset $\cD_t$.

This two-timescale learning scheme enables the model to quickly adapt to time-varying disturbances while 
preserving a stable representation of the underlying dynamics, mitigating oscillations 
and overfitting to recent data. 
Furthermore, the two-timescale scheme reduces the computational cost of online learning compared to standard learning-based MPC methods, as it updates only a subset of parameters at certain control step, which is supported by our experimental results.

The partitioning of parameters $\theta = \{\theta_f, \theta_s \}$ is flexible. For instance, in an $L$-layer neural network, the parameters of the first $L-1$ layers may be grouped into the slow parameter set $\theta_s$, while those of the final layer are grouped into the fast parameter set $\theta_f$.

\subsection{T2S-MPC}

The overall procedure of the proposed Time-Embedded Two-Timescale Neural MPC (T2S-MPC) method is summarized in Algorithm~\ref{alg:online_adaptive_mpc}.
With the learned neural residual model $\hat{\mathbf{r}}_{\theta} (\mathbf{x}(t_i), \mathbf{u}(t_i), \psi(t_i))$ in hand, we can incorporate it with the known nominal model $f$ to estimate the ground truth time-varying dynamics $f_t(\mathbf{x}(t), \mathbf{u}(t))$  as $f(\mathbf{x}(t), \mathbf{u}(t)) + \hat{\mathbf{r}}_{\theta} (\mathbf{x}(t), \mathbf{u}(t), \psi(t))$. Then plugging in 
\begin{align}\label{eq:estimate the model}
    g_t = f(\mathbf{x}(t), \mathbf{u}(t)) + \hat{\mathbf{r}}_{\theta} (\mathbf{x}(t), \mathbf{u}(t), \psi(t))
\end{align}
into the MPC framework in \eqref{eq:mpc_discrete}, it can output the control input $\mathbf{u}_t$ at current time step $t$. This estimated model with the neural residual model is expected to accurately approximate true time-varying dynamics $f_t$, improving dynamic prediction accuracy and the corresponding control solution. 

By combining time-dependent feature with two-timescale learning scheme, the proposed T2S-MPC achieves fast adaptation to non-stationary dynamics while maintaining stable and reliable.

\section{Experiments}
We conduct experiments on the 2D quadrotor system testbed on the $PyBullet$ physics simulator~\citep{coumans2016pybullet}, across two task settings: stabilization and trajectory tracking. To evaluate performance under time-varying dynamics, we introduce diverse disturbance forms, including linearly increasing and periodic disturbances, and compare our proposed method against two baseline methods and two ablation variants. The experiments span a wide range of disturbance magnitudes, periods, and patterns, enabling a comprehensive assessment of both generalization and stability.

\subsection{Task Settings}

\paragraph{System Dynamics}

We consider a 2D quadrotor whose motion is restricted to a vertical plane. The state vector at time $t$ can be directly obtained from the simulation platform, defined as
\begin{equation}
\mathbf{x}(t) = [x(t), \dot{x}(t), z(t), \dot{z}(t), \phi(t), \dot{\phi}(t)]^\top,
\end{equation}
where $x(t)$ and $z(t)$ are the horizontal and vertical positions, $\dot{x}(t)$ and $\dot{z}(t) $ are the corresponding linear velocities, $\phi(t)$ is the pitch angle, and $\dot{\phi}(t)$ is the pitch rate.

\begin{figure}[t]
    \centering

    \begin{subfigure}[t]{0.44\textwidth}
        \centering
        \adjustbox{valign=t, width=\linewidth, height=2.6cm, keepaspectratio}{
            \includegraphics{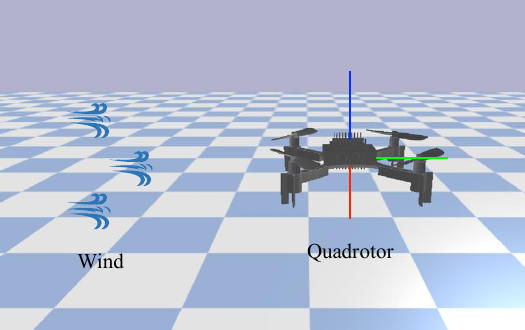}
        }
        \caption{The quadrotor simulation platform.}
        \label{fig:simulation_platform}
    \end{subfigure}
    \hfill
    \begin{subfigure}[t]{0.55\textwidth}
        \centering
        \adjustbox{valign=t, width=\linewidth, height=4.6cm, keepaspectratio}{
            \includegraphics{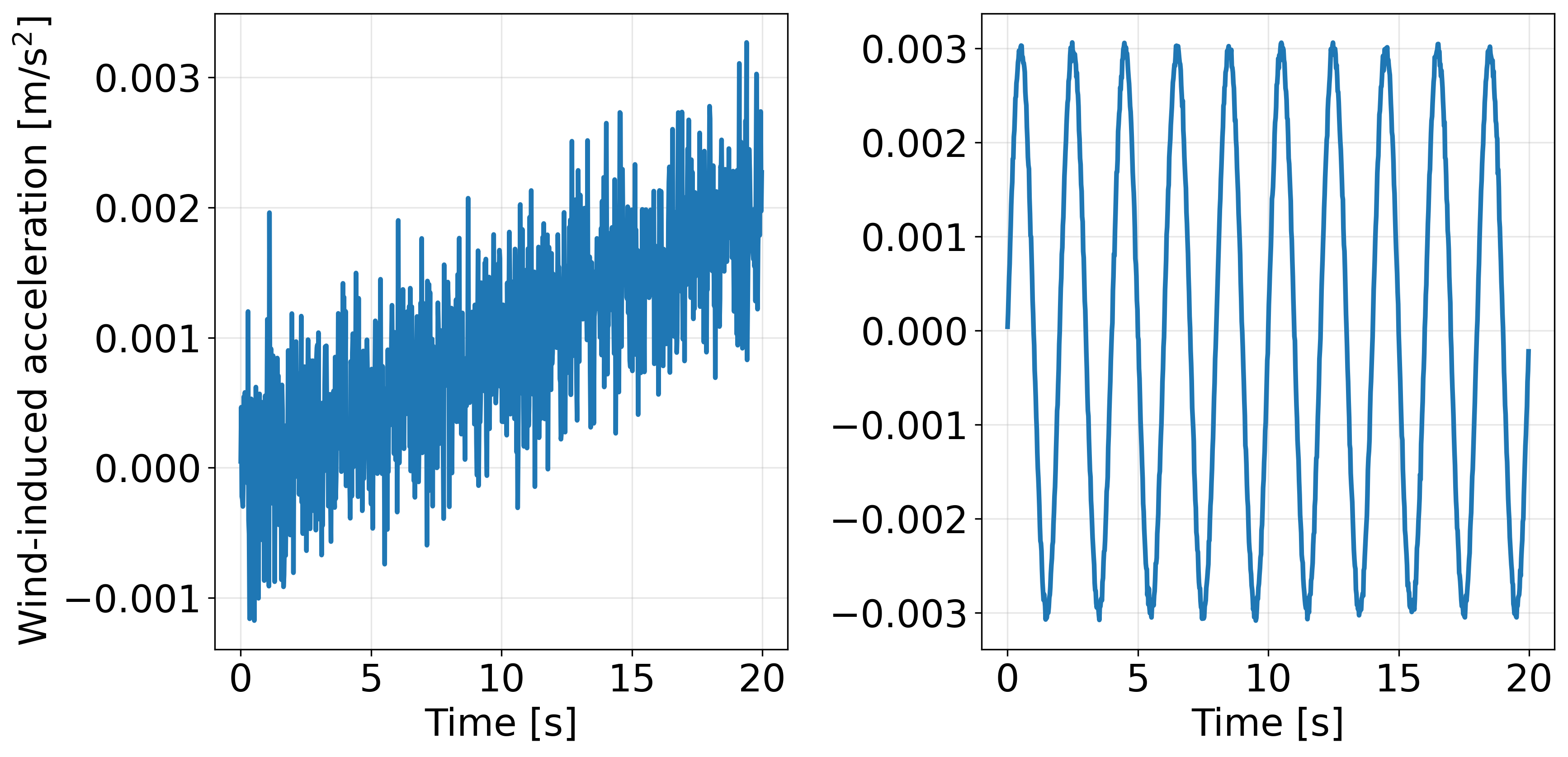}
        }
        \caption{Example of different types of disturbances: drifting (left), periodic (right).}
        \label{fig:disturbance_acceleration_plot}
    \end{subfigure}
    \caption{Simulation platform and disturbance examples.}
    \label{fig:platform_and_disturbances}

\end{figure}

Following~\citep{yuan2022safe}, the dynamics of the 2D quadrotor are given by
\begin{align}
& \ddot{x} = \frac{\sin\phi(t) \big(T_1(t) + T_2(t)\big)}{m_a},  \quad
\ddot{z} = \frac{\cos\phi(t) \big(T_1(t) + T_2(t)\big)}{m_a} - g, \nonumber \\
& \ddot{\phi} = \frac{(T_2(t) - T_1(t))d}{I_{yy}},
\end{align}
where $m_a$ is the mass of the quadrotor, $I_{yy}$ is the moment of inertia about the out-of-plane axis,  $d$ is the distance between the two rotors, and $g$ is the gravitational acceleration. The inputs $T_1(t)$ and $T_2(t)$ denote the thrusts generated by the left and right rotors, respectively. In our experiments, the true system parameters are $m_a = 0.027~\mathrm{kg}$ and $I_{yy} = 1.4 \times 10^{-5}~\mathrm{kg \cdot m^2}$. A visualization of the quadrotor simulation is shown in Fig.~\ref{fig:simulation_platform}.

\paragraph{Tasks:  stabilization and tracking under time-varying disturbance}
In this work, we focus on two standard control tasks~\citep{santos2019adaptive, abdulkareem2022modeling}: (i) stabilization, with the objective to regulate the system to a fixed reference state; and (ii) trajectory tracking, which aims to follow time-varying reference trajectories, including a circular trajectory and a figure-eight trajectory.

To evaluate controller performance under non-stationary environments, we consider two representative classes of time-varying external disturbances $\mathbf{w}(t)$ and add them to the $x$-axis of quadrotor dynamics. These disturbances act on $\ddot{x}$ at the acceleration level indicated as $\mathbf{a}_w(t) \in \mathbb{R}^n$ which are induced by external factors such as wind or load change. Specifically, the two kinds of disturbances are introduced as:
(i) linear drift disturbance, representing a gradual and persistent shift in environment over time; (ii) periodic disturbances, capturing repeated pattern in the real-world. Both disturbances include a stochastic noise term $\epsilon(t)$ and are defined as:

\begin{itemize}
    \item $\mathbf{a}_w(t)_{\mathsf{linear}} = \kappa t + \epsilon(t)$

    \item $\mathbf{a}_w(t)_{\mathsf{period}} = A \sin\left(\frac{2\pi}{T} t\right) + \epsilon(t)$
\end{itemize}
where $\kappa>0 $ is the linear drift rate, and $A$ and $T$ represent the magnitude and the period in the periodic setting. Fig.~\ref{fig:disturbance_acceleration_plot} illustrates the two types of disturbances adopted as the standard experimental setting in our experiments.

\paragraph{Evaluation Metrics}

To quantitatively evaluate controller performance, for each time step $t$, we measure the error as the Euclidean distance between the current position and a desired reference position with position $x_{\mathsf{ref}}$ and $z_{\mathsf{ref}}$ along $x$ and $z$ direction, respectively. Specifically, the error at each time step $t$ is defined as 
\begin{equation}
e(t) = \sqrt{(x(t) - x_{\mathsf{ref}}(t))^2 + (z(t) - z_{\mathsf{ref}}(t))^2},
\end{equation}

For each task with trajectory length $T_\text{sim}$, we report the average error over the simulation horizon:
\begin{equation}\label{eq:error-def}
\bar{e} = \frac{1}{T_\text{sim}} \sum_{t=1}^{T_\text{sim}} e(t).
\end{equation}
For each experiment, 10 independent simulation runs with randomized initial conditions are conducted to account for variability.

\subsection{Baselines and Implementation}

To benchmark the effectiveness of the proposed methods, we compare T2S-MPC with two baselines and two variants, all based on the MPC formulation:
\begin{enumerate}[label=(\roman*)]
    \item \textbf{Nominal MPC}: the classical MPC using only the nominal dynamics model $f$.
    \item \textbf{Neural MPC} \citep{salzmann2023real, mei2025fast}: the neural MPC method that uses a neural network for the residual dynamics model $\hat{\mathbf{r}}_t(\mathbf{x}(t), \mathbf{u}(t))=\hat{\mathbf{r}}_{\theta}(\mathbf{x}(t), \mathbf{u}(t))$, without time embedding $\psi(t)$ input or two-timescale learning scheme.
    \item \textbf{T2S-MPC w/o time emd}: a variant of the proposed T2S-MPC method, without the time embedding. It can also be regarded as Neural MPC with an additional two-timescale updating scheme.
    \item \textbf{T2S-MPC w/o two scales}: another variant of T2S-MPC with time embeddings, but without the two-timescale updating scheme, which is updated under the same pace as Neural MPC.
    \item \textbf{T2S-MPC}: with both time embedding and the two-timescale updating scheme.
\end{enumerate}

For all methods, we use a model predictive control (MPC) framework with a control frequency of 50~Hz as the basis. The MPC optimization problem is formulated with a prediction horizon of $N=20$ steps for $t_p = 0.4$s. 
The running cost function is defined as

\begin{equation}
\ell(\mathbf{x}, \mathbf{u}) = \sum_{k=0}^{N-1} \left( {\mathbf{x}_k}-{\mathbf{x}_{\mathsf{ref},k}})^\top Q ( {\mathbf{x}_k}-{\mathbf{x}_{\mathsf{ref},k}}) + \mathbf{u}_k^\top R \mathbf{u}_k \right),
\end{equation}
where $\mathbf{x}_{\mathsf{ref}} = [x_{\mathsf{ref}}, z_{\mathsf{ref}}]$ is the reference state, $Q = \mathrm{diag}(5, 0.1, 5, 0.1, 0.1, 0.1)$ and $R = \mathrm{diag}(0.1, 0.1)$. 
The terminal cost takes the form of the quadratic state cost.

In this work, we use a multiple $L$-layer neural network, where the parameters of the first $L-1$ layers 
are grouped into the slow parameter set
\begin{equation}
\theta_s = \{W_1,\ldots,W_{L-1}, \mathbf{b}_1,\ldots,\mathbf{b}_{L-1}\},
\end{equation}
while the parameters of the final layer are grouped into the fast parameter set
\begin{equation}
\theta_f = \{W_L, \mathbf{b}_L\}.
\end{equation}
For simplicity, we denote the input as 
$
\mathbf{z}(t)
:=
\big[
\mathbf{x}(t),\;
\mathbf{u}(t),\;
\psi(t)
\big]^\top
\in \mathbb{R}^{n+m+d}$.

The hidden representation produced by the first $N-1$ layers is given by
\begin{equation}
\mathbf{h}(t)
=
\sigma_{L-1}\Big(
W_{L-1}\,
\sigma_{L-2}\big(
\cdots
\sigma_{1}(W_1 \mathbf{z}(t) + \mathbf{b}_1)
\cdots
\big)
+ \mathbf{b}_{L-1}
\Big),
\end{equation}
where $\sigma_i(\cdot)$ denotes the ReLU activation function at layer $i$.
Then, the output of the neural residual model is given by
\begin{equation}
\hat{\mathbf{r}}_{\theta}(\mathbf{x}(t), \mathbf{u}(t), \psi(t))
= W_L \mathbf{h}(t) + \mathbf{b}_L.
\end{equation}

In our experiments, we set $L=3$ and the hidden dimension as 64. The output dimension is 3, and the input dimension is 40, consisting of state-control pair $[\mathbf{x}(t), \mathbf{u}(t)]$ together with a time embedded vector $\psi(t)$ (see Section~\ref{sec:residual-learning}).
The full network therefore contains 6979 trainable parameters. We use Adam optimizers and update the fast-updated parameters $\theta_f$ through~\eqref{eq:fast_update} using initial learning rates $10^{-2}$ after every $t_f=10$ control steps, and update the slow-updated parameters $\theta_s$ through~\eqref{eq:slow_update} using initial learning rates $10^{-3}$ after every $t_s =25$ control steps. The network is trained online using a mean squared error loss~\eqref{eq:loss}.

\subsection{Experimental Results}

\paragraph{Stabilization Performance}
For the stabilization task, the reference state is fixed at $x_{\mathsf{ref}} = [0,\,0,\,1,\,0,\,0,\,0]^\top$. We evaluate performance under two types of disturbances: linearly drifting and periodic settings, illustrated in Fig.~\ref{fig:disturbance_acceleration_plot}.

\begin{figure}[t]
    \centering

    \begin{minipage}{0.48\linewidth}
        \centering
        \includegraphics[width=\linewidth]
        {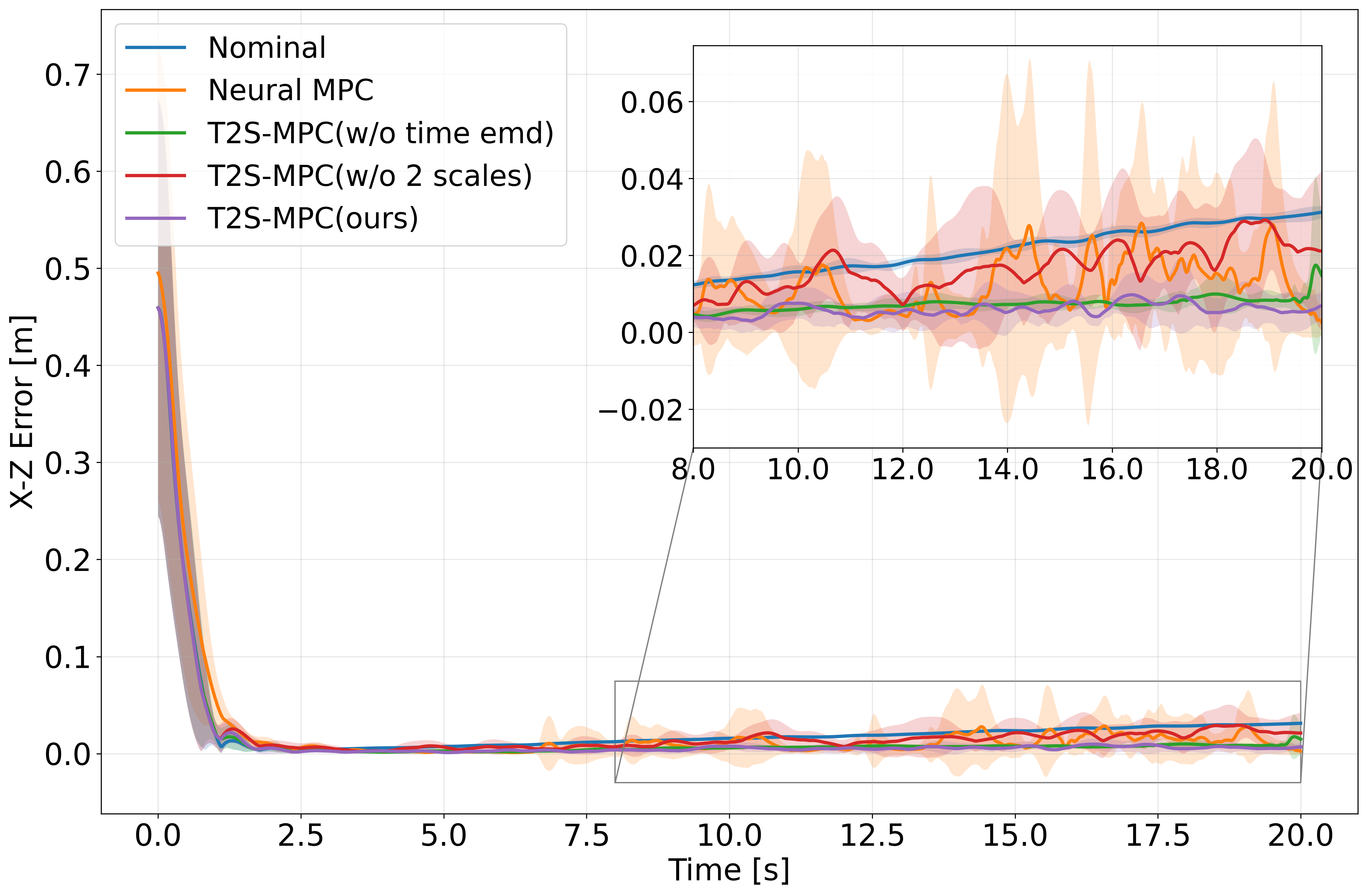}
    \end{minipage}
    \hfill
    \begin{minipage}{0.48\linewidth}
        \centering
        \includegraphics[width=\linewidth]
        {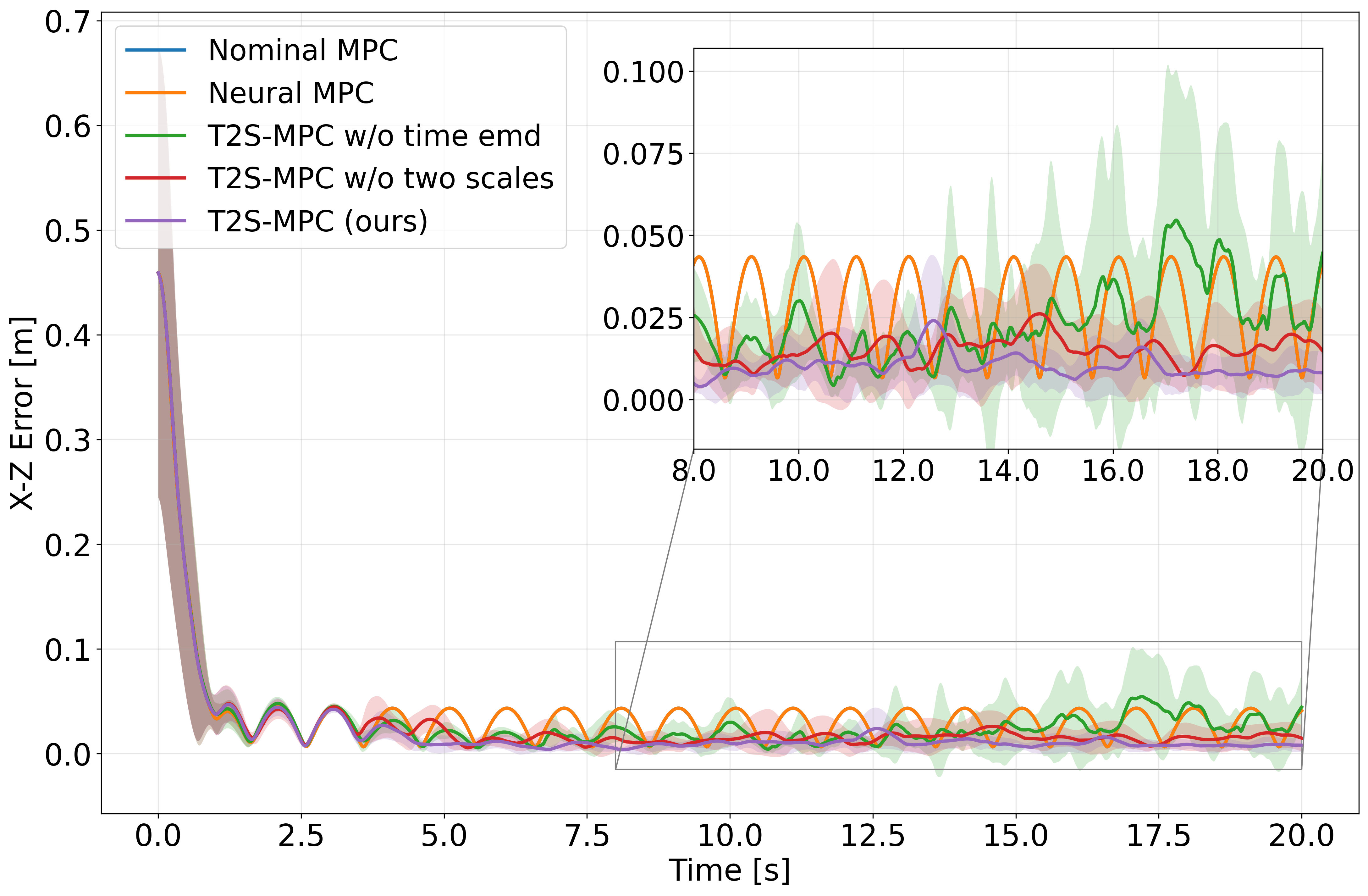}
    \end{minipage}

    \caption{Stabilization error comparison under (left) linearly drifting disturbance and (right) periodic disturbance.}
    \label{fig:stabilization_error_plot}
\end{figure}

Fig.~\ref{fig:stabilization_error_plot} visualize the per-step errors under linear drifting and periodic disturbances. Under linearly drifting disturbances, T2S-MPC achieves superior stabilization performance with significantly reduced oscillations, whereas Nominal MPC and Neural MPC exhibit increasing error over time. Under periodic disturbances, T2S-MPC maintains stable and robust performance, while both baselines suffer from persistent oscillations.

\begin{table}[t]
    \centering
    \caption{Stabilization error under different disturbances. Best performance in \textbf{bold} and second best in \underline{underline}.}
    \vspace{0.5em}
    \label{tab:ablation_study}
    
    \setlength{\tabcolsep}{20pt}
    \renewcommand{\arraystretch}{1.1}
    \small
    
    \begin{tabular}{lcc}
        \toprule
        \textbf{Method} & \textbf{Linearly Drifting} & \textbf{Periodic} \\
        \midrule
        Nominal MPC
        & $0.0261 \pm 0.0058$ 
        & $0.0376 \pm 0.0056$ \\
        
        Neural MPC
        & $0.0221 \pm 0.0069$ 
        & $0.0339 \pm 0.0077$ \\
        
        T2S-MPC w/o time emd 
        & \underline{$0.0161 \pm 0.0059$} 
        & $0.0324 \pm 0.0010$ \\
        
        T2S-MPC w/o two scales 
        & $0.0229 \pm 0.0078$ 
        & \underline{$0.0274 \pm 0.0082$} \\
        
        T2S-MPC (ours)
        & $\mathbf{0.0151 \pm 0.0059}$
        & $\mathbf{0.0228 \pm 0.0066}$ \\
        
        \bottomrule
    \end{tabular}
\end{table}

Table~\ref{tab:ablation_study} summarizes the quantitative control performance across 10 independent runs, measured by the averaged Euclidean error~\eqref{eq:error-def}, for T2S-MPC, its two variants, and two baseline methods. The results show that each component---time embedding and two-timescale learning---independently improves performance, while their combination in T2S-MPC achieves the best overall results, consistently outperforming all alternatives, demonstrating the effectiveness and the complementary roles of these two components.

\paragraph{Trajectory Tracking Performance} In the tracking task, we evaluate performance using two reference trajectories: a circle and a figure-8. The evaluation metrics and disturbance settings are consistent with those used in the stabilization task.

\begin{figure}[t]
    \centering
    \includegraphics[height=6cm]{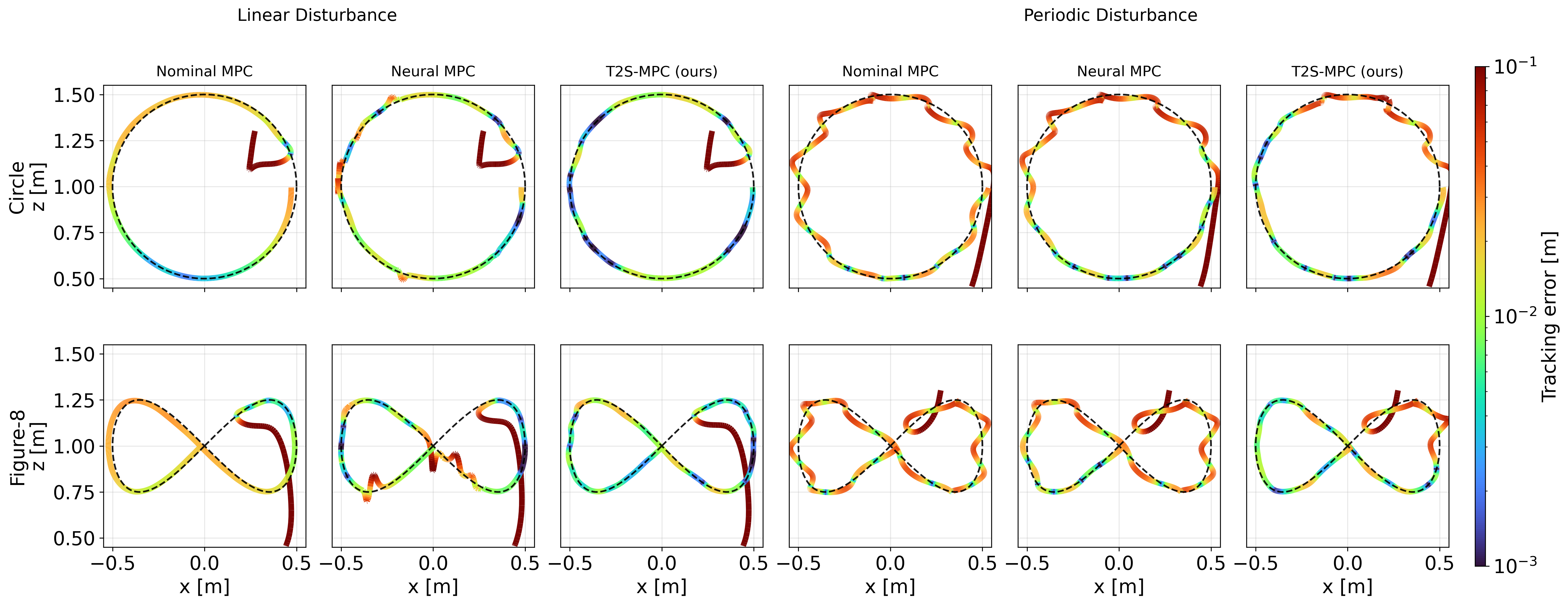}
    \caption{Trajectory tracking performance comparison.}
    \label{fig:tracking}
\end{figure}

Fig.~\ref{fig:tracking} visualizes the tracking error under different trajectory and disturbance settings. It can be observed that the proposed T2S-MPC achieves the best performance, exhibiting accurate tracking with minimal deviation from the reference trajectory.

Table~\ref{tab:tracking} reports the average tracking error, evaluated using the same metric as in the stabilization task. T2S-MPC consistently outperforms all baselines across both settings, demonstrating its robustness to varying trajectories and disturbance conditions.

\begin{table}[t]
    \centering
    \caption{Tracking performance under different disturbances. Best performance in \textbf{bold}.}
    \vspace{0.5em}
    \label{tab:tracking}

    \setlength{\tabcolsep}{20pt}
    \renewcommand{\arraystretch}{1.1}
    \small

    \begin{tabular}{lcc}
        \toprule
        \textbf{Method} & \textbf{Circle} & \textbf{Figure-8} \\
        \midrule

        \multicolumn{3}{c}{\textbf{Linearly Drifting}} \\
        \cmidrule(l{50pt}r{50pt}){1-3}
        Nominal MPC      & $0.0264 \pm 0.0062$ & $0.0263 \pm 0.0058$ \\
        Neural MPC       & $0.0261 \pm 0.0059$ & $0.0228 \pm 0.0073$ \\
        T2S-MPC (ours)   & $\mathbf{0.0216 \pm 0.0064}$ & $\mathbf{0.0199 \pm 0.0062}$ \\

        \midrule

        \multicolumn{3}{c}{\textbf{Periodic}} \\
        \cmidrule(l{50pt}r{50pt}){1-3}
        Nominal MPC      & $0.0408 \pm 0.0058$ & $0.0377 \pm 0.0056$ \\
        Neural MPC       & $0.0404 \pm 0.0079$ & $0.0358 \pm 0.0065$ \\
        T2S-MPC (ours)   & $\mathbf{0.0348 \pm 0.0078}$ & $\mathbf{0.0280 \pm 0.0069}$ \\

        \bottomrule
    \end{tabular}
\end{table}

\paragraph{Disturbance Ablation Study}
To evaluate the robustness and generalization of T2S-MPC, we conduct additional stabilization experiments across a broader range of disturbance conditions, varying the drift form as well as the magnitude and period of the periodic disturbance. As shown in Table~\ref{tab:style}, the proposed method consistently outperforms all baselines across diverse scenarios, particularly under large disturbance magnitudes where stronger compensation is required. Notably, all results are obtained using a single hyperparameter configuration across all conditions without additional tuning, highlighting the method's strong generalization capability.

\begin{table}[t]
    \centering
    \caption{Stabilization performance under more different disturbances.}
    \vspace{0.5em}
    \label{tab:style}

    \setlength{\tabcolsep}{2pt}
    \renewcommand{\arraystretch}{1.15}
    \small

    % \resizebox{0.9\textwidth}{!}{%
    \begin{tabular}{lcccccccc}
        \toprule

        & \multicolumn{6}{c}{\textbf{Periodic Disturbance}} 
        & \multicolumn{2}{c}{\textbf{Drifting Disturbance}} \\

        \cmidrule(lr){2-7} \cmidrule(lr){8-9}

        \textbf{Magnitude}
        & \multicolumn{2}{c}{0.001}
        & \multicolumn{2}{c}{0.003}
        & \multicolumn{2}{c}{0.005}
        & \multicolumn{2}{c}{\textbf{Drift Forms}} \\

        \cmidrule(lr){2-3}
        \cmidrule(lr){4-5}
        \cmidrule(lr){6-7}
        \cmidrule(lr){8-9}

        \textbf{Period}
        & 2 & 4
        & 2 & 4
        & 2 & 4
        & Polynomial & Linear with Step\\

        \midrule

        Nominal MPC
        & 0.0194 & 0.0201
        & 0.0376 & 0.0395
        & 0.0562 & 0.0620
        & 0.0294 & 0.0375 \\

        Neural MPC
        & 0.0209 & 0.0210
        & 0.0339 & 0.0405
        & 0.0418 & 0.0653
        & 0.0231 & 0.0266 \\

        T2S-MPC (ours)
        & \textbf{0.0163} & \textbf{0.0191}
        & \textbf{0.0228} & \textbf{0.0322}
        & \textbf{0.0297} & \textbf{0.0453}
        & \textbf{0.0179} & \textbf{0.0204} \\

        \bottomrule
      \end{tabular}%
    % }
\end{table}

\paragraph{Computation Efficiency Analysis}
We also evaluate computational efficiency, focusing on neural network updates as the primary computational bottleneck in learning-based MPC. Under a consistent hardware setup (a laptop with an Intel Core i7-8550U processor and 8GB RAM), the average computation time is 1027.5~ms for the fast update and 1021.9~ms for the slow update, whereas Neural MPC requires 1268.2~ms per update. These results demonstrate that the decoupled update scheme in T2S-MPC is comparable and even more efficient than updating all parameters simultaneously, although the proposed model has a larger parameter count due to the additional time embedding. This efficiency highlights strong potential for real-time implementation.

\FloatBarrier
\section{Conclusion}
In this work, we address the control problem of partially known time-varying dynamical systems by adopting online neural network-based model learning within the controller. We decompose the dynamics into a known nominal model and a residual component, and propose T2S-MPC, which learns the residual model online using a neural network. Compared to prior neural residual approaches, we explicitly handle time-varying dynamics by incorporating a structured time embedding as input to the neural network, providing explicit temporal information, together with a two-timescale update scheme. This enables the controller to capture nonstationary dynamics while balancing rapid adaptation and stable learning. We integrate the learned residual with the nominal model within the MPC controller and evaluate T2S-MPC on quadrotor stabilization and tracking tasks, where it outperforms four MPC-based methods. Furthermore, T2S-MPC demonstrates robust and generalizable performance under diverse time-varying disturbances. Future work will focus on improving robustness to high-frequency disturbances, which remains a broader challenge in learning-based control for diverse control systems and real-world robotic platforms.

\bibliographystyle{apalike}
\bibliography{reference}

@article{hewing2020learning,
  title={Learning-based model predictive control: Toward safe learning in control},
  author={Hewing, Lukas and Wabersich, Kim P and Menner, Marcel and Zeilinger, Melanie N},
  journal={Annual Review of Control, Robotics, and Autonomous Systems},
  volume={3},
  number={1},
  pages={269--296},
  year={2020},
  publisher={Annual Reviews}
}

@book{ioannou1996robust,
  title={Robust adaptive control},
  author={Ioannou, Petros A and Sun, Jing},
  volume={1},
  year={1996},
  publisher={PTR Prentice-Hall Upper Saddle River, NJ}
}

@incollection{aastrom1995adaptive,
  title={Adaptive control},
  author={{\AA}str{\"o}m, Karl Johan},
  booktitle={Mathematical System Theory: The Influence of RE Kalman},
  pages={437--450},
  year={1995},
  publisher={Springer}
}

@article{kabzan2019learning,
  title={Learning-based model predictive control for autonomous racing},
  author={Kabzan, Juraj and Hewing, Lukas and Liniger, Alexander and Zeilinger, Melanie N},
  journal={IEEE Robotics and Automation Letters},
  volume={4},
  number={4},
  pages={3363--3370},
  year={2019},
  publisher={IEEE}
}

@article{ren2022tutorial,
  title={A tutorial review of neural network modeling approaches for model predictive control},
  author={Ren, Yi Ming and Alhajeri, Mohammed S and Luo, Junwei and Chen, Scarlett and Abdullah, Fahim and Wu, Zhe and Christofides, Panagiotis D},
  journal={Computers \& Chemical Engineering},
  volume={165},
  pages={107956},
  year={2022},
  publisher={Elsevier}
}

@article{mei2025fast,
  title={Fast Online Adaptive Neural MPC via Meta-Learning},
  author={Mei, Yu and Zhou, Xinyu and Yu, Shuyang and Srivastava, Vaibhav and Tan, Xiaobo},
  journal={IFAC-PapersOnLine},
  volume={59},
  number={30},
  pages={377--382},
  year={2025},
  publisher={Elsevier}
}

@article{kouvaritakis2016model,
  title={Model predictive control},
  author={Kouvaritakis, Basil and Cannon, Mark},
  journal={Switzerland: Springer International Publishing},
  volume={38},
  number={13-56},
  pages={7},
  year={2016},
  publisher={Springer}
}

@article{garcia1989model,
  title={Model predictive control: Theory and practice—A survey},
  author={Garcia, Carlos E and Prett, David M and Morari, Manfred},
  journal={Automatica},
  volume={25},
  number={3},
  pages={335--348},
  year={1989},
  publisher={Elsevier}
}

@inproceedings{jiahao2023online,
  title={Online dynamics learning for predictive control with an application to aerial robots},
  author={Jiahao, Tom Z and Chee, Kong Yao and Hsieh, M Ani},
  booktitle={Conference on Robot Learning},
  pages={2251--2261},
  year={2023},
  organization={PMLR}
}

@inproceedings{saveriano2017data,
  title={Data-efficient control policy search using residual dynamics learning},
  author={Saveriano, Matteo and Yin, Yuchao and Falco, Pietro and Lee, Dongheui},
  booktitle={2017 IEEE/RSJ International Conference on Intelligent Robots and Systems (IROS)},
  pages={4709--4715},
  year={2017},
  organization={IEEE}
}

@article{bock1984multiple,
  title={A multiple shooting algorithm for direct solution of optimal control problems},
  author={Bock, Hans Georg and Plitt, Karl-Josef},
  journal={IFAC Proceedings Volumes},
  volume={17},
  number={2},
  pages={1603--1608},
  year={1984},
  publisher={Elsevier}
}

@article{sun2021online,
  title={Online learning of unknown dynamics for model-based controllers in legged locomotion},
  author={Sun, Yu and Ubellacker, Wyatt L and Ma, Wen-Loong and Zhang, Xiang and Wang, Changhao and Csomay-Shanklin, Noel V and Tomizuka, Masayoshi and Sreenath, Koushil and Ames, Aaron D},
  journal={IEEE Robotics and Automation Letters},
  volume={6},
  number={4},
  pages={8442--8449},
  year={2021},
  publisher={IEEE}
}

@article{smith2020online,
  title={Online simultaneous semi-parametric dynamics model learning},
  author={Smith, Joshua and Mistry, Michael},
  journal={IEEE Robotics and Automation Letters},
  volume={5},
  number={2},
  pages={2039--2046},
  year={2020},
  publisher={IEEE}
}

@article{williams2019locally,
  title={Locally weighted regression pseudo-rehearsal for online learning of vehicle dynamics},
  author={Williams, Grady and Goldfain, Brian and Rehg, James M and Theodorou, Evangelos A},
  journal={arXiv preprint arXiv:1905.05162},
  year={2019}
}

@article{yuan2022safe,
  title={Safe-control-gym: A unified benchmark suite for safe learning-based control and reinforcement learning in robotics},
  author={Yuan, Zhaocong and Hall, Adam W and Zhou, Siqi and Brunke, Lukas and Greeff, Melissa and Panerati, Jacopo and Schoellig, Angela P},
  journal={IEEE Robotics and Automation Letters},
  volume={7},
  number={4},
  pages={11142--11149},
  year={2022},
  publisher={IEEE}
}

@article{chakrabarty2023meta,
  title={Meta-learning of neural state-space models using data from similar systems},
  author={Chakrabarty, Ankush and Wichern, Gordon and Laughman, Christopher R},
  journal={IFAC-PapersOnLine},
  volume={56},
  number={2},
  pages={1490--1495},
  year={2023},
  publisher={Elsevier}
}

@inproceedings{yan2024mpc,
  title={MPC of uncertain nonlinear systems with meta-learning for fast adaptation of neural predictive models},
  author={Yan, Jiaqi and Chakrabarty, Ankush and Rupenyan, Alisa and Lygeros, John},
  booktitle={2024 IEEE 20th International Conference on Automation Science and Engineering (CASE)},
  pages={1910--1915},
  year={2024},
  organization={IEEE}
}

@article{muthirayan2025meta,
  title={Meta-learning online control for linear dynamical systems},
  author={Muthirayan, Deepan and Kalathil, Dileep and Khargonekar, Pramod P},
  journal={IEEE Transactions on Automatic Control},
  volume={70},
  number={6},
  pages={4163--4169},
  year={2025},
  publisher={IEEE}
}

@article{o2022neural,
  title={Neural-fly enables rapid learning for agile flight in strong winds},
  author={O’Connell, Michael and Shi, Guanya and Shi, Xichen and Azizzadenesheli, Kamyar and Anandkumar, Anima and Yue, Yisong and Chung, Soon-Jo},
  journal={Science Robotics},
  volume={7},
  number={66},
  pages={eabm6597},
  year={2022},
  publisher={American Association for the Advancement of Science}
}

@article{wang2024tutorial,
  title={A tutorial on Gaussian process learning-based model predictive control},
  author={Wang, Jie and Zhang, Youmin},
  journal={arXiv preprint arXiv:2404.03689},
  year={2024}
}

@article{chowdhary2014bayesian,
  title={Bayesian nonparametric adaptive control using Gaussian processes},
  author={Chowdhary, Girish and Kingravi, Hassan A and How, Jonathan P and Vela, Patricio A},
  journal={IEEE transactions on neural networks and learning systems},
  volume={26},
  number={3},
  pages={537--550},
  year={2014},
  publisher={IEEE}
}

@article{chee2022knode,
  title={KNODE-MPC: A knowledge-based data-driven predictive control framework for aerial robots},
  author={Chee, Kong Yao and Jiahao, Tom Z and Hsieh, M Ani},
  journal={IEEE Robotics and Automation Letters},
  volume={7},
  number={2},
  pages={2819--2826},
  year={2022},
  publisher={IEEE}
}

@article{salzmann2023real,
  title={Real-time neural MPC: Deep learning model predictive control for quadrotors and agile robotic platforms},
  author={Salzmann, Tim and Kaufmann, Elia and Arrizabalaga, Jon and Pavone, Marco and Scaramuzza, Davide and Ryll, Markus},
  journal={IEEE Robotics and Automation Letters},
  volume={8},
  number={4},
  pages={2397--2404},
  year={2023},
  publisher={IEEE}
}

@inproceedings{salzmann2024learning,
  title={Learning for casadi: Data-driven models in numerical optimization},
  author={Salzmann, Tim and Arrizabalaga, Jon and Andersson, Joel and Pavone, Marco and Ryll, Markus},
  booktitle={6th Annual Learning for Dynamics \& Control Conference},
  pages={541--553},
  year={2024},
  organization={PMLR}
}

@article{santos2019adaptive,
  title={An adaptive dynamic controller for quadrotor to perform trajectory tracking tasks},
  author={Santos, Milton Cesar Paes and Rosales, Claudio Dar{\'\i}o and Sarapura, Jorge Antonio and Sarcinelli-Filho, M{\'a}rio and Carelli, Ricardo},
  journal={Journal of Intelligent \& Robotic Systems},
  volume={93},
  number={1},
  pages={5--16},
  year={2019},
  publisher={Springer}
}

@article{abdulkareem2022modeling,
  title={Modeling and nonlinear control of a quadcopter for stabilization and trajectory tracking},
  author={Abdulkareem, Ademola and Oguntosin, Victoria and Popoola, Olawale M and Idowu, Ademola A},
  journal={Journal of Engineering},
  volume={2022},
  number={1},
  pages={2449901},
  year={2022},
  publisher={Wiley Online Library}
}

@article{zhang2016understanding,
  title={Understanding deep learning requires rethinking generalization},
  author={Zhang, Chiyuan and Bengio, Samy and Hardt, Moritz and Recht, Benjamin and Vinyals, Oriol},
  journal={arXiv preprint arXiv:1611.03530},
  year={2016}
}

@article{Wu2019RealTimeMLMPC,
  author  = {Wu, Zhe and Rincon, David and Christofides, Panagiotis D.},
  title   = {Real-time adaptive machine-learning-based predictive control of nonlinear processes},
  journal = {Industrial \& Engineering Chemistry Research},
  year    = {2019},
  volume  = {59},
  number  = {6},
  pages   = {2275--2290},
  doi     = {10.1021/acs.iecr.9b03055}
}

@misc{coumans2016pybullet,
  title={Pybullet, a python module for physics simulation for games, robotics and machine learning},
  author={Coumans, Erwin and Bai, Yunfei},
  year={2016}
}

@article{chen2021adaptive,
  title={Adaptive control for systems with time-varying parameters—A survey},
  author={Chen, Kaiwen and Astolfi, Alessandro},
  journal={Trends in Nonlinear and Adaptive Control: A Tribute to Laurent Praly for His 65th Birthday},
  pages={217--247},
  year={2021},
  publisher={Springer}
}

@inproceedings{gradu2023adaptive,
  title={Adaptive regret for control of time-varying dynamics},
  author={Gradu, Paula and Hazan, Elad and Minasyan, Edgar},
  booktitle={Learning for dynamics and control Conference},
  pages={560--572},
  year={2023},
  organization={PMLR}
}

@article{wu2025tutorial,
  title={A tutorial review of machine learning-based model predictive control methods},
  author={Wu, Zhe and Christofides, Panagiotis D and Wu, Wanlu and Wang, Yujia and Abdullah, Fahim and Alnajdi, Aisha and Kadakia, Yash},
  journal={Reviews in Chemical Engineering},
  volume={41},
  number={4},
  pages={359--400},
  year={2025},
  publisher={De Gruyter}
}

@article{wu2019real,
  title={Real-time adaptive machine-learning-based predictive control of nonlinear processes},
  author={Wu, Zhe and Rincon, David and Christofides, Panagiotis D},
  journal={Industrial \& Engineering Chemistry Research},
  volume={59},
  number={6},
  pages={2275--2290},
  year={2019},
  publisher={ACS Publications}
}

@article{richards2021adaptive,
  title={Adaptive-control-oriented meta-learning for nonlinear systems},
  author={Richards, Spencer M and Azizan, Navid and Slotine, Jean-Jacques and Pavone, Marco},
  journal={arXiv preprint arXiv:2103.04490},
  year={2021}
}

@article{shi2021meta,
  title={Meta-adaptive nonlinear control: Theory and algorithms},
  author={Shi, Guanya and Azizzadenesheli, Kamyar and O'Connell, Michael and Chung, Soon-Jo and Yue, Yisong},
  journal={Advances in Neural Information Processing Systems},
  volume={34},
  pages={10013--10025},
  year={2021}
}

@article{minasyan2021online,
  title={Online control of unknown time-varying dynamical systems},
  author={Minasyan, Edgar and Gradu, Paula and Simchowitz, Max and Hazan, Elad},
  journal={Advances in Neural Information Processing Systems},
  volume={34},
  pages={15934--15945},
  year={2021}
}

@inproceedings{chowdhary2010concurrent,
  title={Concurrent learning for convergence in adaptive control without persistency of excitation},
  author={Chowdhary, Girish and Johnson, Eric},
  booktitle={49th IEEE Conference on Decision and Control (CDC)},
  pages={3674--3679},
  year={2010},
  organization={IEEE}
}

@article{parikh2019integral,
  title={Integral concurrent learning: Adaptive control with parameter convergence using finite excitation},
  author={Parikh, Anup and Kamalapurkar, Rushikesh and Dixon, Warren E},
  journal={International Journal of Adaptive Control and Signal Processing},
  volume={33},
  number={12},
  pages={1775--1787},
  year={2019},
  publisher={Wiley Online Library}
}

@inproceedings{glushchenko2021drem,
  title={I-DREM MRAC with time-varying adaptation rate \& no a priori knowledge of control input matrix sign to relax PE condition},
  author={Glushchenko, Anton I and others},
  booktitle={2021 European Control Conference (ECC)},
  pages={2175--2180},
  year={2021},
  organization={IEEE}
}

@article{goel2024composite,
  title={Composite adaptive control for time-varying systems with dual adaptation},
  author={Goel, Raghavv and Roy, Sayan Basu},
  journal={IEEE Transactions on Automatic Control},
  volume={70},
  number={1},
  pages={487--494},
  year={2024},
  publisher={IEEE}
}

@article{patil2022adaptive,
  title={Adaptive control of time-varying parameter systems with asymptotic tracking},
  author={Patil, Omkar Sudhir and Sun, Runhan and Bhasin, Shubhendu and Dixon, Warren E},
  journal={IEEE Transactions on Automatic Control},
  volume={67},
  number={9},
  pages={4809--4815},
  year={2022},
  publisher={IEEE}
}

@article{zhang2024global,
  title={Global asymptotic fault-tolerant tracking for time-varying nonlinear complex systems with prescribed performance},
  author={Zhang, Zhikai and Dong, Yi and Duan, Guangren},
  journal={Automatica},
  volume={159},
  pages={111345},
  year={2024},
  publisher={Elsevier}
}

@article{shi2022prescribed,
  title={Prescribed-time asymptotic tracking control of strict feedback systems with time-varying parameters and unknown control direction},
  author={Shi, Wenrui and Hou, Mingzhe and Duan, Guangren},
  journal={IEEE Transactions on Circuits and Systems I: Regular Papers},
  volume={69},
  number={12},
  pages={5259--5272},
  year={2022},
  publisher={IEEE}
}

@article{kumpati1990identification,
  title={Identification and control of dynamical systems using neural networks},
  author={Kumpati, S Narendra and Kannan, Parthasarathy and others},
  journal={IEEE Transactions on neural networks},
  volume={1},
  number={1},
  pages={4--27},
  year={1990}
}

@article{hunt1992neural,
  title={Neural networks for control systems—a survey},
  author={Hunt, Kenneth J and Sbarbaro, Daniel and {\.Z}bikowski, R and Gawthrop, Peter J},
  journal={Automatica},
  volume={28},
  number={6},
  pages={1083--1112},
  year={1992},
  publisher={Elsevier}
}

@article{achterhold2024learning,
  title={Learning a Terrain-and Robot-Aware Dynamics Model for Autonomous Mobile Robot Navigation},
  author={Achterhold, Jan and Guttikonda, Suresh and Kreber, Jens U and Li, Haolong and Stueckler, Joerg},
  journal={arXiv preprint arXiv:2409.11452},
  year={2024}
}

@article{hanover2021performance,
  title={Performance, precision, and payloads: Adaptive nonlinear mpc for quadrotors},
  author={Hanover, Drew and Foehn, Philipp and Sun, Sihao and Kaufmann, Elia and Scaramuzza, Davide},
  journal={IEEE Robotics and Automation Letters},
  volume={7},
  number={2},
  pages={690--697},
  year={2021},
  publisher={IEEE}
}

@article{langeron2015modeling,
  title={A modeling framework for deteriorating control system and predictive maintenance of actuators},
  author={Langeron, Yves and Grall, Antoine and Barros, Anne},
  journal={Reliability Engineering \& System Safety},
  volume={140},
  pages={22--36},
  year={2015},
  publisher={Elsevier}
}

%%%%%%%%%%%%%%%%%%%%%%%%%%%%%%%%%%%%%%%%%%%%%%%%%%%%%%%%%%%%

\end{document}